\documentclass[conference]{IEEEtran}
\makeatletter
\def\ps@IEEEtitlepagestyle{%
  \def\@evenfoot{}%
}
\makeatother
\IEEEoverridecommandlockouts
\usepackage{iftex}
\ifXeTeX\usepackage{fontspec}\else\usepackage[T1]{fontenc}\usepackage[utf8]{inputenc}\usepackage{newtxtext}\fi
\usepackage{cite}
\usepackage{amsmath,amssymb,amsfonts}
\usepackage{graphicx}
\usepackage{textcomp}
\usepackage{booktabs}
\usepackage{adjustbox}
\usepackage{url}
\begin{document}
\title{Synthetic Twins: Closing the Gap Between TTS Test Sets and Real Recordings in Turkish ASR Evaluation}

\author{\IEEEauthorblockN{Nur Yağmur Tuncer}
\IEEEauthorblockA{\textit{Data Science and Innovation} \\
\textit{Ata Technology Platforms}\\
Istanbul, Turkey \\
yagmur.tuncer@atptech.com}
\and
\IEEEauthorblockN{Mustafa Sertaç Türkel}
\IEEEauthorblockA{\textit{Data Science and Innovation} \\
\textit{Ata Technology Platforms}\\
Istanbul, Turkey \\
sertac.turkel@atptech.com}
\and
\IEEEauthorblockN{Fatma Nur Korkmaz}
\IEEEauthorblockA{\textit{Data Science and Innovation} \\
\textit{Ata Technology Platforms}\\
Istanbul, Turkey \\
fatmanur.korkmaz@atptech.com}
\and
\IEEEauthorblockN{Ahmet Tuğrul Bayrak}
\IEEEauthorblockA{\textit{Data Science and Innovation} \\
\textit{Ata Technology Platforms}\\
Istanbul, Turkey \\
tugrul.bayrak@atptech.com}
}

\maketitle
{\let\thefootnote\relax\footnotetext{Accepted at ISAS 2026. \copyright~2026 IEEE. Personal use of this material is permitted. Permission from IEEE must be obtained for all other uses, in any current or future media, including reprinting/republishing this material for advertising or promotional purposes, creating new collective works, for resale or redistribution to servers or lists, or reuse of any copyrighted component of this work in other works.}}
\begin{abstract}
Speech synthesized by text-to-speech engines is a low-cost substitute for recorded test sets in automatic speech recognition, yet an off-the-shelf synthetic test set is worth only a few dozen real recordings for estimating the real error rate of a recognizer. This paper examines which properties of real recordings are missing from synthetic speech and how much of the gap can be closed. For 507 Turkish Common Voice recordings, \textit{synthetic twins} are built: a ladder of synthetic copies that adds, one factor at a time, the pitch of the speaker, a clone of the same speaker, the real recording channel and the speaking rate. At each step, the agreement between ASR behaviour on the copy and on the real recording is measured. Across three zero-shot cloning engines and panels of six and 19 recognizers, the speaker clone is the step that improves agreement on which sentences are hard, raising the Spearman correlation from .41 to .55 for Chatterbox. Matching the channel or adding generic noise corrects the error level when the synthetic speech is easier than real speech. The twin with the highest worth is equivalent to 70 to 122 real recordings, depending on the recognizer panel, compared with about 9 for a stock voice. The findings replicate on VoxForge, a second corpus in which speaker identity is known rather than inferred: the clone raises sentence agreement from .31 to .50 and the worth from 8 to about 50 recordings. Agreement on which sentences are hard is distinct from agreement on how hard they are. The paper concludes with a practical recipe and a reporting checklist for synthetic ASR test sets.
\end{abstract}

\section{Introduction}
Evaluating automatic speech recognition (ASR) requires recorded test speech, which is expensive to collect, depends on the consent of the speakers and is scarce for most languages. Text-to-speech (TTS) offers an alternative: any text can be read aloud at low cost, so test sets can be generated for new domains, rare phenomena or new languages~\cite{b1,b2}. The reliability of such test sets is less clear. Synthetic speech raises false alarms, i.e.\ failures that do not occur on human speech, in 21\% to 34\% of cases~\cite{b2}, and models that perform almost perfectly on synthetic audio can perform poorly on real speech~\cite{b3}.

Prior work mainly establishes that a gap exists. The present paper quantifies it as the number of real recordings that a synthetic test set is worth and asks \textit{why} synthetic speech is less informative than real speech, that is, which properties of real recordings it lacks, and how the gap can be closed. The paper makes three contributions. First, it introduces \textit{synthetic twins}, a ladder of controlled interventions that brings each synthetic utterance progressively closer to its real counterpart in speaker pitch, speaker identity, recording channel and timing; the ladder is evaluated at four levels of fidelity and by the number of real recordings that the synthetic set is worth. Second, using three cloning engines, panels of 6 and 19 recognizers, a generic-noise control and a second corpus with known speakers, it shows that matching the speaker reproduces part of the sentence-level difficulty pattern, while channel and noise shift the error level. Third, it proposes a practical recipe and a reporting checklist for ASR evaluation on synthetic speech.

\section{Related Work}
CrossASR~\cite{b1} uses TTS to generate test cases for differential testing, and Lau et al.~\cite{b2} show that many of the failures found in this way are false alarms and train a classifier to filter them. 2M-BELEBELE~\cite{b5} finds that three ASR models are ranked unreliably when human FLEURS recordings are replaced by an MMS-TTS rendering, even for high-resource languages. WildASR~\cite{b3} avoids TTS after observing an almost sixfold increase in errors from synthetic to real child speech. For Turkish, Tuncer et al.~\cite{rat} show that synthetic sets from nine TTS engines preserve recognizer rankings but are worth at most 31 real recordings. These works establish the gap but do not decompose it.

Minixhofer et al.~\cite{b6} quantify the distance between synthetic and real speech with utterance-level speaker, prosody and environment statistics and reduce this distance by about 10\% by conditioning the TTS on the true values. Rossenbach et al.~\cite{b7} examine how the choice of TTS model shapes synthetic training data for ASR. Neither study asks whether a fixed recognizer behaves in the same way on a synthetic copy of a test set, a property referred to here as \textit{evaluation} fidelity and measured per sentence. Closest to the present ladder, Lebryk et al.~\cite{b8} clone the voices of benchmark speakers and find that model behaviour on the clones resembles behaviour on the real recordings more closely than behaviour on generic voices. Their goal is to detect benchmark-specific optimization, and they match neither channel nor timing and do not measure sentence-level agreement. For Turkish clinical ASR, Tuncer et al.~\cite{tms} train on synthetic speech and confirm the gains on real recordings.

Prediction-powered inference~\cite{b9,b10} combines a small set of gold labels with many inexpensive predictions and quantifies their value as an effective sample size. The same perspective is adopted here to express what a synthetic test set is worth in real recordings.

\section{Data and Measures}
The experiments use 743 utterances from the Turkish test split of Common Voice 17~\cite{b11}, drawn with a fixed seed as part of the balanced multi-source Turkish benchmark TurkASR-Bench~\cite{tab}. Contributors record themselves on their own devices, so channels differ widely between utterances. Section~IV repeats part of the twin experiment on 507 read utterances from the Turkish section of VoxForge~\cite{b13}, in which speaker identity is known.

%The reference text of each utterance is read by three zero-shot engines, XTTS-v2~\cite{b14}, Chatterbox~\cite{b15} and VoxCPM2~\cite{b16}, both with built-in voices and as clones of the real speaker. Each synthetic utterance keeps the identifier of its real counterpart, so every comparison is paired by sentence.

The reference text of each utterance is read by three zero-shot engines, XTTS-v2~\cite{b14}, Chatterbox~\cite{b15} and VoxCPM2~\cite{b16}, both with built-in voices and as clones of the real speaker. Each synthetic utterance keeps the identifier of its real counterpart, so every comparison is paired by sentence. The synthetic test set\footnote{\url{https://huggingface.co/datasets/yagmurtuncer/turkasr-bench-tts}} and the synthetic twins\footnote{\url{https://huggingface.co/datasets/yagmurtuncer/turkasr-bench-twins}} are publicly available.

All audio is transcribed by 19 open recognizers: eight sizes of the Whisper family~\cite{b17} up to large-v3-turbo, three Turkish Whisper fine-tunes, five Turkish wav2vec2 and XLS-R models~\cite{b18}, MMS-1B-all~\cite{b19}, Qwen3-ASR-1.7B~\cite{b20} and SeamlessM4T v2~\cite{b21}. The twin experiments use a representative panel of six of them, namely Whisper tiny, medium and large-v3, XLS-R 300M fine-tuned on Common Voice 8, MMS-1B-all and Qwen3-ASR-1.7B, and all 19 as a robustness check. Before scoring, references and hypotheses are normalized to Unicode NFC, lower-cased with Turkish rules and stripped of punctuation, and digits are spelled out. Corpus-level WER is reported, where lower values are better.

\subsection{Four Levels of Fidelity}
A synthetic test set can be adequate for one use and inadequate for another. Agreement with real speech is therefore measured at four levels, and a pass threshold was fixed for each level before any result was computed:

\begin{itemize}\item \textit{Ranking:} Kendall's $\tau$~\cite{b22} between the WERs of the recognizers on the two versions, together with the fraction of significantly different recognizer pairs whose order is preserved. The threshold is $\tau \ge .80$ with at least 90\% of such pairs.
\item \textit{Absolute error:} The signed gap $\Delta = \mathrm{WER}_{\mathrm{syn}} - \mathrm{WER}_{\mathrm{real}}$ of each recognizer, with paired bootstrap intervals. The threshold is $|\Delta| \le 2$ points for at least 80\% of the recognizers.
\item \textit{Sentence difficulty:} The Spearman correlation $\rho$ between real and synthetic per-sentence WER, averaged over recognizers, and the false-alarm rate, i.e.\ the fraction of recognizer--sentence cells with an error only on the synthetic version. The threshold is $\rho \ge .60$ with at most 20\% false alarms.
\item \textit{Calibration:} A linear regression of real on synthetic WER, fitted in a leave-one-recognizer-out manner and scored by its mean absolute error on the held-out recognizer. The threshold is 3 points.\end{itemize}

Intervals are based on 2,000 paired bootstrap resamples of sentences, and a pass decision requires the bootstrap failure probability to remain below .05 after Holm--Bonferroni correction across engines and levels.

\subsection{What Is a Synthetic Test Set Worth?}
Fidelity is also expressed as a single number, the \textit{worth} of a synthetic set $S$. It is defined as the number $n$ of randomly drawn real recordings for which $E|\hat{W}(n) - W| = |W(S) - W|$, where $W$ is the WER of a recognizer on all real recordings and $\hat{W}(n)$ its WER on the $n$ drawn recordings, averaged over recognizers. A set worth 50 conveys as much about real WER as 50 random recordings, so the worth is the effective sample size of the set in the sense of prediction-powered inference~\cite{b9}. Because the 507 recordings of the experiment are selected by speaker match rather than drawn at random, their WER deviates from $W$, so even the same real recordings are worth fewer than 507 recordings.

\section{Synthetic Twins}
Each synthetic utterance is built as a progressively closer \textit{twin} of its real counterpart by changing one factor at a time with XTTS-v2, which supports zero-shot voice cloning:

\begin{itemize}\item \textit{R0 (stock voice):} One of four built-in voices of the engine.
\item \textit{R1 (pitch-matched voice):} The built-in voice whose median F0 is closest to that of the real speaker, on the same side of a split at 165~Hz.
\item \textit{R2 (speaker clone):} A clone of the real speaker conditioned on a \textit{different} utterance by the same speaker.
\item \textit{R3 (+ channel):} R2 passed through the channel of the real recording. The channel consists of long-term spectrum equalization with one-third-octave smoothing limited to $\pm$20~dB, the background noise of the real recording taken from its quietest non-silent frames at the original level, loudness matching and an MP3 round trip at 64~kbit/s, as for the Common Voice originals.
\item \textit{R4 (+ speaking rate):} R3 with the clone re-synthesized at the speaking rate of the real recording, using the speed control of XTTS-v2 with the ratio of speech durations clipped to the range from 0.6 to 1.8.
\item \textit{R0 + channel:} The channel of R3 applied to the stock voice, which separates speaker effects from channel effects.
\item \textit{Oracle clone:} A clone conditioned on the target utterance itself, which gives an upper bound for cloning.\end{itemize}

Since Common Voice does not expose speaker identities in the split used here, all 9,650 test utterances are embedded with an ECAPA-TDNN speaker encoder~\cite{b23,b24}, and for each target the most similar utterance with a different sentence is selected. Cloning with XTTS and matching with ECAPA rely on different encoders, so the match is not self-confirming. Where both utterances carry a gender label, the labels agree in all 265 cases. Of the 743 targets, 513 have a match with a cosine similarity of at least 0.60, and the 507 of these with a pitch estimate form the experiment, transcribed by the six representative recognizers of Section~III. The twins resemble their speakers: the mean ECAPA similarity to the real recording is .55 for clones and .61 for oracle clones, compared with .12 for stock voices, .14 for pitch-matched voices and .70 between two real recordings of the same speaker.

\begin{table*}[t]\centering\footnotesize\setlength{\tabcolsep}{4pt}\caption{Synthetic-twin ladder with XTTS-v2}
\begin{adjustbox}{max width=\textwidth}\begin{tabular}{lccccc}\toprule
Condition & Sentence $\rho$ ($\uparrow$) & False alarms ($\downarrow$) & $\Delta$ pts & |$\Delta$| pts ($\downarrow$) & Worth ($\uparrow$) \\ \midrule
R0 stock voice & .38 [.30, .46] & 24\% & $-$5.7 & 5.8 & 9 \\
R1 pitch-matched voice & .38 [.30, .46] & 23\% & $-$7.9 & 7.9 & 5 \\
R0 + channel & .35 [.26, .44] & 24\% & $-$1.8 & 1.9 & 45 \\
R2 speaker clone & .44 [.35, .52] & 25\% & +0.5 & 3.6 & 32 \\
R3 clone + channel & .44 [.35, .52] & 26\% & +5.2 & 5.2 & 19 \\
R4 clone + channel + rate & .41 [.33, .49] & 29\% & +13.5 & 13.5 & 3 \\
Oracle clone (target utterance) & .51 [.43, .58] & 24\% & +0.5 & 3.5 & 32 \\
Same real recordings & 1 & --- & 0.0 & 0.0 & 251 \\
\bottomrule\end{tabular}\end{adjustbox}\end{table*}
\begin{figure*}[t]\centering\includegraphics[width=\textwidth]{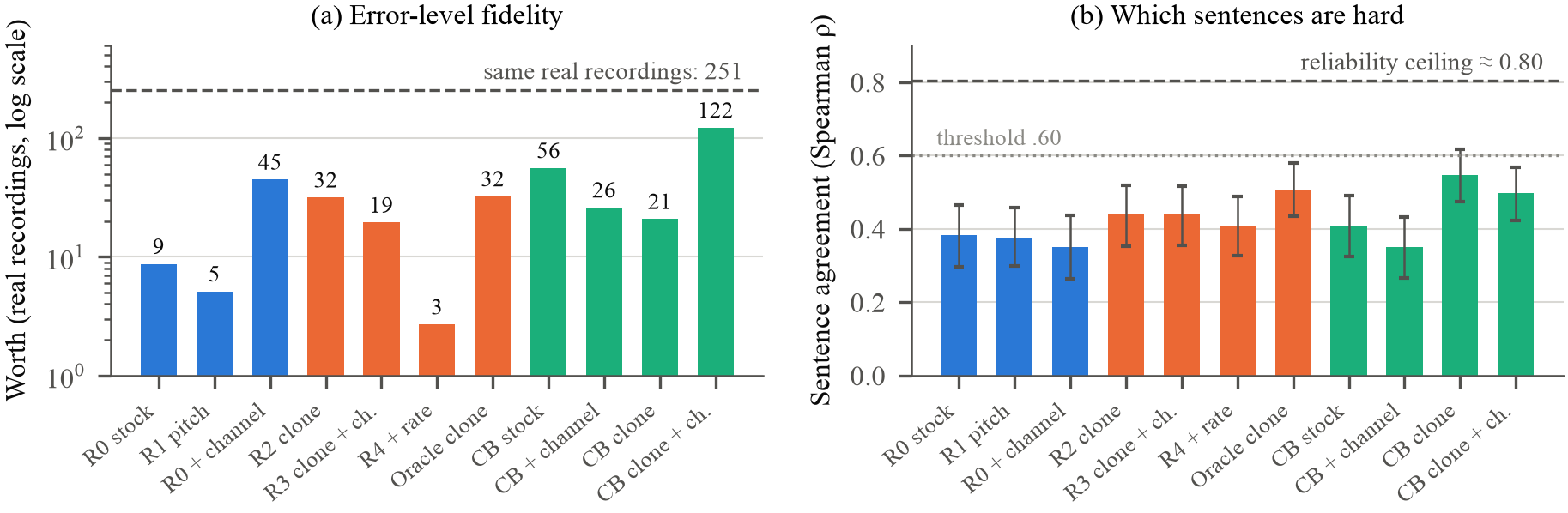}\caption{Worth and sentence agreement along the twin ladder}\end{figure*}
Table~I and Fig.~1 show the ladder for XTTS-v2. In Table~I, $\Delta$ is the mean signed difference between synthetic and real WER, negative when the synthetic speech is easier, and the last row, the real recordings of the same sentences, sets the ceiling for worth. In Fig.~1, panel~(a) shows the worth on a logarithmic scale and panel~(b) the sentence agreement with 95\% bootstrap intervals; blue, orange and green bars denote XTTS stock voices, XTTS clones and Chatterbox (CB). The dashed lines mark the worth of the same real recordings and the reliability ceiling of the six-recognizer difficulty measure, and the dotted line marks the pass threshold. Table~II tests the steps against each other; its intervals are 95\% intervals from 2,000 bootstrap resamples over sentences, and its $p$ values are Holm-corrected within each column and engine.

Matching the channel corrects the error level of a voice that is easier to recognize than real speech. The stock voice of XTTS is easier than real speech, with $\Delta = -5.7$ points. The channel of the real recording reduces the gap from 5.8 to 1.9 points, a significant reduction of 3.9 points, and raises the worth from 9 to 45 recordings, while sentence agreement remains essentially unchanged. Matching the speaker is more effective than matching pitch alone: a clone of the speaker also narrows the gap significantly, by 2.2 points, whereas a voice that only matches the pitch of the speaker widens it by 2.1 points. The XTTS clone raises sentence agreement from .38 to .44, a gain in the expected direction that does not reach significance after correction, and an oracle clone conditioned on the target recording itself raises it significantly, by 0.12. The effects of speaker and channel are not additive. A clone already inherits part of the channel of its reference recording, so applying the target channel in addition in R3 makes the synthetic speech harder than real speech by a further 1.7 points.

The highest worth among the XTTS twin sets is 32 random recordings, while the same 507 real recordings are worth 251, and the oracle clone reaches $\rho = .51$, below the threshold of .60. Matching the speaking rate in R4 makes the synthetic speech considerably harder and raises $\Delta$ to 13.5 points. Since intelligibility was not measured separately, this increase cannot be attributed to timing itself and may reflect a limitation of the speed control of XTTS-v2, as discussed in Section~V.

\begin{table*}[t]\centering\footnotesize\setlength{\tabcolsep}{4pt}\caption{Paired differences between ladder steps}
\begin{adjustbox}{max width=\textwidth}\begin{tabular}{lcccc}\toprule
Step & $\Delta$ sentence $\rho$ ($\uparrow$) & p (Holm) & $\Delta$ |gap| pts ($\downarrow$) & p (Holm) \\ \midrule
R0 stock $\to$ R1 pitch & $-$0.007 [$-$0.078, +0.061] & 1.000 & +2.1 [+1.0, +3.1] & 0.005 \\
R0 stock $\to$ R0 + channel & $-$0.033 [$-$0.078, +0.013] & 0.625 & $-$3.9 [$-$4.5, $-$2.8] & < .001 \\
R0 stock $\to$ R2 clone & +0.055 [$-$0.022, +0.132] & 0.625 & $-$2.2 [$-$4.2, $-$0.7] & 0.016 \\
R1 pitch $\to$ R2 clone & +0.062 [$-$0.015, +0.138] & 0.625 & $-$4.3 [$-$6.1, $-$2.5] & < .001 \\
R2 clone $\to$ R3 clone + channel & $-$0.000 [$-$0.050, +0.050] & 1.000 & +1.7 [+0.5, +3.0] & 0.016 \\
R2 clone $\to$ Oracle clone & +0.069 [+0.001, +0.137] & 0.246 & $-$0.1 [$-$1.0, +0.9] & 0.867 \\
R0 stock $\to$ Oracle clone & +0.124 [+0.046, +0.208] & 0.021 & $-$2.3 [$-$4.3, $-$0.8] & 0.016 \\
Chatterbox: R0 stock $\to$ R0 + channel & $-$0.056 [$-$0.107, $-$0.006] & 0.060 & +2.5 [+0.2, +3.9] & 0.111 \\
Chatterbox: R0 stock $\to$ R2 clone & +0.141 [+0.055, +0.221] & < .001 & +0.8 [$-$1.2, +2.6] & 0.434 \\
Chatterbox: R2 clone $\to$ R3 clone + channel & $-$0.050 [$-$0.095, $-$0.006] & 0.060 & $-$2.0 [$-$3.6, +0.2] & 0.188 \\
\bottomrule\end{tabular}\end{adjustbox}\end{table*}
The four central steps are repeated with a second cloning engine, Chatterbox~\cite{b15}, conditioned on the same reference recordings. Table~III reports the results on Common Voice (CV) and VoxForge (VF), each with 507 sentences and the six representative recognizers; Kendall's $\tau$ over the six WERs equals 1.00 for every CV condition and is therefore shown for VF only. The Chatterbox clones resemble the speakers more closely than the XTTS clones, with an ECAPA similarity of .59 compared with .55, and the results refine the XTTS findings. The clone now improves sentence agreement significantly, from .41 to .55, a gain of 0.14 at $p < .001$. The closer Chatterbox clones and the XTTS oracle clones, with similarities of .59 and .61, both improve agreement significantly, which suggests that a speaker clone carries part of the difficulty of a sentence when it is sufficiently close to the speaker. Unlike that of XTTS, the stock voice of Chatterbox is already close to real error levels, with a worth of 56, and adding the channel makes it harder than real speech and lowers the worth to 26. In both engines, the channel improves the match where synthetic speech is easier than real speech, a pattern that the third engine confirms. Combining a Chatterbox clone, which is easier than real speech, with the real channel gives the twin with the highest worth in the study, 122 recordings, with a 95\% confidence interval from 49 to 239 and a gap of 1.4 points. The same real recordings reach a worth of 251 with an interval from 66 to 446, so worth estimates based on six recognizers should be read together with their intervals.

\begin{table*}[t]\centering\footnotesize\setlength{\tabcolsep}{4pt}\caption{Chatterbox twins on Common Voice and VoxForge}
\begin{adjustbox}{max width=\textwidth}\begin{tabular}{lccccccc}\toprule
Condition (Chatterbox) & CV $\rho$ ($\uparrow$) & CV |$\Delta$| ($\downarrow$) & CV worth ($\uparrow$) & VF $\rho$ ($\uparrow$) & VF |$\Delta$| ($\downarrow$) & VF $\tau$ ($\uparrow$) & VF worth ($\uparrow$) \\ \midrule
R0 stock voice & .41 & 2.7 & 56 & .31 & 4.5 & .73 & 8 \\
R0 + channel & .35 & 5.1 & 26 & .35 & 5.5 & .73 & 5 \\
R2 speaker clone & .55 & 3.4 & 21 & .50 & 2.9 & .73 & 52 \\
R3 clone + channel & .50 & 1.4 & 122 & .53 & 1.8 & .87 & 54 \\
Same real recordings & 1 & 0.0 & 251 & 1 & 0.0 & 1 & 156 \\
\bottomrule\end{tabular}\end{adjustbox}\end{table*}
As a third cloning engine, VoxCPM2~\cite{b16}, a recent zero-shot engine, is conditioned on the same reference recordings and their transcripts. Its clones are the closest to the speakers among the three engines, with an ECAPA similarity of .65 compared with .12 for its stock voice. VoxCPM2 confirms the clone effect, raising sentence agreement significantly from .50 to .62, and reaches the highest sentence agreement of the study. Its speech is, however, considerably easier than real speech: with the six representative recognizers, the WER on its clones is 10.2 points lower than on the real recordings, which limits the worth of the clones to 3 recordings. Consistent with the channel pattern, the real channel narrows this gap by 3.3 points to 6.9 points and raises the worth to 7. An engine can thus capture \textit{which} sentences are hard without capturing \textit{how} hard they are, so sentence agreement and error level should be checked separately.

Sentence difficulty measured with six recognizers has limited reliability. Repeating the analysis with all 19 recognizers raises the split-half reliability on real speech from .78 to .91. Table~IV compares the three engines under both panels. The conclusions remain unchanged. The gain of the clone in sentence agreement remains significant for Chatterbox, at 0.16, and for VoxCPM2, at 0.09, both with $p < .001$, and for XTTS it rises to 0.08. With 19 recognizers, the twins with the highest worth are the Chatterbox clone with channel, at 70 recordings, and the XTTS clone, at 60, compared with 10 for the stock XTTS voice and 309 for the real recordings. The large gain of these twins over the stock XTTS voice is stable across both panels.

\begin{table*}[t]\centering\footnotesize\setlength{\tabcolsep}{4pt}\caption{Three cloning engines with 6 and 19 recognizers}
\begin{adjustbox}{max width=\textwidth}\begin{tabular}{lcccccc}\toprule
Engine $\cdot$ condition & $\rho$, 6 rec. ($\uparrow$) & $\rho$, 19 rec. ($\uparrow$) & |$\Delta$|, 19 ($\downarrow$) & $\tau$, 19 ($\uparrow$) & Worth, 6 ($\uparrow$) & Worth, 19 ($\uparrow$) \\ \midrule
XTTS $\cdot$ stock voice & .38 & .40 & 5.3 & .80 & 9 & 10 \\
XTTS $\cdot$ stock + channel & .35 & .37 & 2.2 & .86 & 45 & 43 \\
XTTS $\cdot$ speaker clone & .44 & .48 & 2.5 & .91 & 32 & 60 \\
XTTS $\cdot$ clone + channel & .44 & .47 & 4.9 & .90 & 19 & 22 \\
Chatterbox $\cdot$ stock voice & .41 & .47 & 2.5 & .90 & 56 & 66 \\
Chatterbox $\cdot$ stock + channel & .35 & .41 & 5.1 & .90 & 26 & 21 \\
Chatterbox $\cdot$ speaker clone & .55 & .62 & 3.2 & .95 & 21 & 23 \\
Chatterbox $\cdot$ clone + channel & .50 & .58 & 1.9 & .91 & 122 & 70 \\
VoxCPM2 $\cdot$ stock voice & .50 & .56 & 13.4 & .78 & 2 & 2 \\
VoxCPM2 $\cdot$ stock + channel & .51 & .55 & 11.5 & .80 & 2 & 3 \\
VoxCPM2 $\cdot$ speaker clone & .62 & .65 & 9.3 & .85 & 3 & 4 \\
VoxCPM2 $\cdot$ clone + channel & .57 & .61 & 6.5 & .84 & 7 & 8 \\
Same real recordings & 1 & 1 & 0.0 & 1 & 251 & 309 \\
\bottomrule\end{tabular}\end{adjustbox}\end{table*}
A generic alternative to the real channel is to add noise or reverberation from a standard collection, namely MUSAN background noise~\cite{musan} at a fixed signal-to-noise ratio and real room impulse responses~\cite{rir}. Applied to the Piper engine~\cite{piper}, whose speech is much easier than real speech with $\Delta = -8.6$ points, such augmentation also brings the error level close to the real one and raises the worth from 4 to between 31 and 39, but it lowers sentence agreement from .48 to between .39 and .41. On XTTS, noise at a signal-to-noise ratio of 20~dB raises the worth from 9 to 15, compared with 45 for the real channel. Generic degradation can thus adjust the error \textit{level}, while the speaker clone is the step that improves sentence-level agreement.

Speaker matches in Common Voice are inferred, and its channel includes MP3 coding. The Chatterbox steps are therefore repeated on VoxForge~\cite{b13}, a corpus of Turkish read speech recorded by volunteers in sessions of about ten prompts each. A session identifies the speaker, so each clone is conditioned on the recording from the \textit{same session} with a different prompt that is most similar to the target under ECAPA, with a median similarity of .80, and no similarity threshold is needed. Of the 743 VoxForge utterances drawn for the multi-source benchmark, 507 utterances from 165 sessions are used, matching the size of the Common Voice experiment; 315 of them were selected in identifier order and 192 at random, and results on the random subset are reported separately. The audio is uncompressed WAV, so the channel step omits the MP3 round trip. The clones are again close to the speakers, with an ECAPA similarity to the real recording of .67, compared with .12 for the stock voice and .79 between two real recordings.

Table~III compares both corpora and shows that the speaker effect replicates: the clone raises sentence agreement from .31 to .50, a significant gain of 0.19 at $p < .001$, now with speaker identities that are known rather than inferred. On the 192 randomly drawn sentences alone, the gain is the same, from .34 to .51, which confirms that the partly ordered selection does not account for the result. The channel pattern also replicates. On VoxForge, the stock voice is \textit{harder} than real speech by 4.5 points, and the channel widens the gap by a further 1.0 point, whereas the clone is easier than real speech by 2.3 points and the channel narrows the gap by 1.1 points. The clones are worth 52 recordings without the channel and 54 with it, with 95\% confidence intervals from 30 to 76 and from 26 to 95, compared with 8 for the stock voice, whose interval ranges from 3 to 18. On VoxForge, $\tau$ rises from .73 for the stock voice to .87 for the clone with channel; with six recognizers, a single swapped pair of close systems corresponds to a difference of .13 in $\tau$.

Re-synthesizing the oracle twin with a different random seed and scoring both copies gives a sentence agreement of $\rho = .69$ between the two synthetic versions. The agreement between real recordings and twins, between .51 and .55, thus reaches 74\% to 80\% of the agreement of a twin with a second rendering of itself, so part of the remaining gap reflects the variability of zero-shot TTS rather than a missing property of real speech.

Sentence difficulty is itself measured with error. When the six recognizers are split into halves, the Spearman--Brown reliability of per-sentence WER is .78 on real speech and .83 on oracle twins, which limits the observable agreement to about .80. After correction for this attenuation, agreement rises from .48 for stock voices to .55 for clones and .63 for oracle clones.

Per sentence, the difference between real and oracle-twin WER shows no relation to the speaking rate of the real recording, its background level, its share of pauses or the speaker similarity of the clone, with correlations of .06, $-$.07, $-$.03 and $-$.01, respectively. Only sentence length shows a weak association, with $\rho = .18$. Clear reading deviations, in which the two strongest recognizers transcribe the same non-reference words on the real recording but the reference on the twin, occur in only 5 of 507 sentences. The gap is therefore not concentrated in a few recordings with noise, fast speech or misreadings but is spread over many sentences, which is consistent with fine-grained differences in how a person articulates a particular sentence. Given the width of the intervals in Table~II, ladder studies of this kind require several hundred sentences.

\section{Discussion}
Synthetic speech supports coarse comparisons of recognizers, but it does not separate close systems, report absolute accuracy or locate failures. Stating the worth of the set makes this scope explicit: a set equivalent to ten recordings cannot support a WER reported to one decimal place.

If a synthetic set is to replace real recordings, the speakers of a small real sample should be cloned, and both the choice of cloning engine and the use of the channel should be decided with the worth measure on this sample: the audio is passed through the real recording channel only if the synthetic speech is still easier than real speech. In the experiments of this paper, this procedure raised the worth of the synthetic set from about 10 to between 70 and 122 recordings on Common Voice, depending on the recognizer panel, and from 8 to 54 recordings on VoxForge. A small real anchor set is used both for this decision and for reporting the worth of the synthetic set.

When ASR results are reported on synthetic speech, the following information should accompany them: the TTS engine and voices and whether speakers were cloned, a small in-domain set of real recordings, ranking agreement with an interval, the absolute WER difference per system, sentence-level agreement and the false-alarm rate, and the worth of the synthetic set in real recordings. % [2026-11-23] kod paylasilmayacak: "small tool ... released together with the code" cumlesi kaldirildi

In Common Voice, the speaker of the reference recording is found by embedding similarity; on VoxForge, where recording sessions identify the speaker, the clone effect is reproduced, which indicates that it does not arise from this matching procedure. The speaking-rate step is bounded by what current TTS speed control can render intelligibly, and a duration model that preserves intelligibility would allow timing to be studied on its own. The XLS-R recognizer was fine-tuned on Common Voice, which may favour it on the Common Voice recordings; this does not apply to VoxForge. Voices were cloned only to measure evaluation fidelity; the clones are available only on request, and their reference recordings come from public corpora, which are distributed under the CC0 and GPL licences. % [2026-11-23] "not released" -> "available only on request" (HF onayli erisim); kod paylasilmayacagi icin "the code reproduces them" kaldirildi

The ladder separates factors that previous studies have not examined separately. The gap that remains after matching voice and channel is partly due to TTS variability and may partly reflect properties of individual readings, such as hesitations, emphasis and local speaking rate, that the recording-level measures of Section~IV do not capture and that current zero-shot TTS does not reproduce from a reference recording. Across the three engines, clones that resemble the speaker more closely also reproduce more of the sentence difficulty.

\section{Conclusion}
Off-the-shelf TTS voices are worth only a few dozen real recordings at most. Synthetic twins make it possible to measure this gap and to reduce it. Cloning the speakers and, where the synthetic speech is easier than real speech, matching the recording channel raises the worth of a synthetic set seven- to fourteen-fold, from 9 to 122 recordings with six recognizers and from 10 to 70 with 19, and clones that closely resemble the speaker reproduce part of the difficulty of each sentence. Part of the remaining gap is due to TTS variability, and the rest may reflect the particular way in which a person reads a sentence. 

Future work will extend the ladder to conversational speech and further languages, add a duration model to study timing in isolation, and determine the minimum size of a real anchor set that still selects the right twin configuration.

\end{document}